\documentclass[runningheads,a4paper]{llncs}

\usepackage{enumerate,amsmath,amssymb,graphicx}
\usepackage{url}
\usepackage{hyperref} 
\usepackage{fancyvrb}
\usepackage{epsfig}
\usepackage{tikz,xspace,multicol}
\usepackage{enumitem}
\usepackage{algorithm}

\newcommand{\CASE}[1]{\STATE \textbf{case} #1\textbf{:} \begin{ALC@g}}
\newcommand{\ENDCASE}{\end{ALC@g}}

\newcommand{\DEFAULT}{\STATE \textbf{default:} \begin{ALC@g}}
\newcommand{\ENDDEFAULT}{\end{ALC@g}}
\newcommand{\DEFAULTLINE}[1]{\STATE \textbf{default:} }

\usetikzlibrary{positioning,shapes,arrows,automata}
\tikzset{ state/.style={draw,ellipse,initial text=} }

\tikzset{
  loop above right/.style={above right, out= 60, in= 30, loop},
  loop above left/.style ={above left,  out=150, in=120, loop},
  loop below right/.style={below right, out=330, in=300, loop},
  loop below left/.style ={below left,  out=240, in=210, loop}
}

\newcommand{\rpatl}	{{\sf PATL+R}\xspace}

\newcommand{\ie}    	{i.e., }

\newcommand{\wrt}   	{w.r.t.\ }
\newcommand{\true}   	{{\mathtt{true}}}

\newcommand{\dist}   	{{\sf{Dist}}}

\newcommand{\rew}   	{{\sf{rew}}}

\def\F{\Diamond} 
\def\P{\mathbf{P}}
\def\R{\mathbf{R}}
\def\D{\mathbf{D}}
\def\U{\mathbf{U}}

\def\X{\bigcirc}

\def\E{\mathcal{E}}

\newcommand{\HIST}[2]{\mathsf{Hist}_{#2}(#1)}
\newcommand{\HISTSTR}[3]{\mathsf{Hist}_{#2}^{#3}(#1)}

\newcommand{\PLAN}[3]{\mathsf{Plan}^{#3}_{#2}(#1)}

\newcommand{\ACT}{\mathsf{Act}}
\newcommand{\AG}{\mathsf{Ag}}
\newcommand{\ASP}{\mathsf{Ap}}
\newcommand{\BR}{\mathbf{BR}}
\newcommand{\payoff}{\wp}

\newcommand{\SAT}{\mathsf{Sat}}

\newcommand{\CAR} {\mathsf{CAR}}
\newcommand{\CPR} {\mathsf{CPR}}

\newcommand{\catch} {\texttt{\scriptsize Catch}}
\newcommand{\iskip} {\texttt{\scriptsize Skip}}

\newcommand{\caught} {\textit{\footnotesize caught}}
\newcommand{\collision} {\textit{\footnotesize collision}}
\newcommand{\dropped} {\textit{\footnotesize dropped}}
\newcommand{\score} {\textit{\footnotesize score}}

\newcommand{\FS}{\rightarrow}
\newcommand{\CAL}[1]{\mathcal{#1}}
\newcommand{\EEE}{\mathbb{E}}
\newcommand{\DDD}{\mathbb{D}}
\newcommand{\FFF}{\CAL{F}}
\newcommand{\LLL}{\CAL{L}}
\newcommand{\MMM}{\CAL{M}}
\newcommand{\VVV}{\CAL{V}}
\newcommand{\ER}{\CAL{R}}

\newcommand{\PPP}{\mathbb{P}}

\newcommand{\RRR}{\mathbb{R}}

\newcommand{\TRANS}[1]{\stackrel{#1}{\longrightarrow}}
\newenvironment{GRAMMAR}{\[\begin{array}{lcl}}{\end{array}\]}
\newcommand{\VERTICAL}{\  \mid\hspace{-3.0pt}\mid \ }

\definecolor{blue}{rgb}{0,0,1}

\urldef{\mailsa}\path|chunyan.mu@abdn.ac.uk,m.najib@hw.ac.uk,n.oren@abdn.ac.uk|

\renewcommand{\Game}{\mathcal{G}}
\newcommand{\psmas}{PSMAS\xspace}

\usepackage{algorithm}
\usepackage[noend]{algpseudocode}
\usepackage{amsmath}
\usepackage{thm-restate}
\usepackage{cleveref}

\begin{document}

\title{Responsibility-aware Strategic Reasoning in \\ Probabilistic Multi-Agent Systems}
\titlerunning{Responsibility-aware Strategic Reasoning in  Probabilistic MASs}

 \author{Chunyan Mu\inst{1} \and Muhammad Najib\inst{2} \and Nir Oren\inst{1}} 
 \authorrunning{C. Mu \and M. Najib \and N. Oren} 
 \institute{$^1$ Department of Computing Science, University of Aberdeen, U.K.\\
 	$^2$ Department of Computer Science, Heriot-Watt University, U.K. \\
 	\mailsa
 }
 
\maketitle

\begin{abstract}

Responsibility plays a key role in the development and deployment of trustworthy autonomous systems.
In this paper, we focus on the problem of strategic reasoning in probabilistic multi-agent systems with responsibility-aware agents. We introduce \rpatl, a variant of Probabilistic Alternating-time Temporal Logic. \rpatl's novelty lies in its incorporation of modalities for causal responsibility, providing a framework for responsibility-aware multi-agent strategic reasoning.  
We present an approach to synthesise joint strategies that satisfy an outcome specified in \rpatl while optimising the share of expected causal responsibility and reward. This provides a notion of balanced distribution of responsibility and reward gain among agents. To this end, we utilise the Nash equilibrium as the solution concept for our strategic reasoning problem and demonstrate how to compute responsibility-aware Nash equilibrium strategies via a reduction to parametric model checking of concurrent stochastic multi-player games.
\end{abstract}

\section{Introduction}
\label{sec:intro}

Strategic decision-making and reasoning in multi-agent systems (MASs) operating in a dynamic and uncertain environment is a challenging research problem. Formalisms such as ATL~\cite{AlurHK02} and PATL~\cite{ChenL07} provide important frameworks for reasoning about strategic decision making in MASs. 
These formalisms have also found notable applications in the field of multi-agent planning~\cite{van2002tractable,jamroga2004strategic,schnoor2010strategic,demri2023model,chrpa2024verifying}. 
Recently there has been a surge in research focused on the trustworthiness of autonomous systems, highlighting the importance of responsibility as a crucial element of AI systems~\cite{kaur2022trustworthy,dignum2020responsibility}. Consequently, the notion of a \textit{responsibility-aware agent} has emerged~\cite{yazdanpanah2023reasoning,kobayashi2023formal,cosner2023learning}. Such an agent must  consider not only its own reward but also its responsibility with respect to the tasks assigned to it.

This paper explores the problem of strategic reasoning where agents are aware of \emph{causal responsibility} \cite{ChocklerH04,ParkerGL23}, which captures the impact of an agent's or coalition's actions, or inaction, on outcomes. 
Analysing causal responsibility enables a balanced distribution of rewards or penalties among agents based on their contribution to outcomes\footnote{In particular, when combined with \textit{norms} \cite{HenneOBKB21}, although we emphasise that the study of normative systems is beyond the scope of this paper.}. This provides a way for ensuring fairness among agents. For example, consider a scenario where two agents are involved in carrying out a \textit{plan} (i.e., a joint strategy) assigned by a principal. If both agents share the same degree of responsibility for the execution of this plan, they should receive a fair share of reward (e.g., each receives the same amount of money). Moreover, agents may pursue different plans based on their preferences, meaning that individual agents may deviate from an overarching plan. Inspired by the concept of \textit{equilibria for multi-agent planning}~\cite{bowling2003formalization}, we utilise the Nash equilibrium (NE) as the solution concept to compute stable plans. If a plan is a NE, then no agent can benefit from unilaterally deviating from the plan.

\paragraph{Contributions.}
This paper focuses on the strategic reasoning problem with \textit{step-bounded temporal properties}, where agents have access to \textit{randomised memoryless} strategies and seek to optimise the balance between responsibility and reward (i.e., agents are responsibility-aware).
We present a parametric model that captures both the uncertainty and strategic interactions in multi-agent systems of such a setting. We then introduce the \rpatl logic as a language for specifying properties that encompass responsibility and temporal objectives. By utilising the approach from the parametric model checking paradigm, we demonstrate how to model check \rpatl formulae against our proposed model. We show that such a model checking problem can be solved in PSPACE. 
We further demonstrate how we can utilise the parametric model to compute NE strategies/plans with respect to agents' \textit{utility functions} that consider both reward and responsibility. 
These NE strategies/plans are in a sense optimal for each agent, considering trade-offs between responsibility and reward. We show that such a computation can also be done in PSPACE. Omitted proofs can be found in supplementary material.

\paragraph{Related work.} 
Previous research has explored the incorporation of responsibility into automated planning.
The work of Alechina et al. \cite{AlechinaHL17} formalises the attribution of responsibility (or blame) for joint plan failures. Our work diverges from this by focusing on how the degree of responsibility can actively influence plan generation rather than post-hoc analysis of a predetermined plan.
Parker et al.'s work \cite{ParkerGL23} is more closely related to ours, as it considers causal responsibility for plan selection. Specifically, they consider both \textit{Causal Active Responsibility} and \textit{Causal Passive Responsibility}. Our concept of responsibility is based on theirs. The primary difference between their work and ours lies in the model used. They consider an imperfect information setting (we use a perfect information setting), and they only consider a deterministic setting, while ours is probabilistic.
The work of Baier et al. \cite{Baier0M21} bears similarity to ours in the sense that they consider games and strategies, unlike the more traditional model in classical planning. However, unlike our work, none of these works consider game-theoretic (specifically, NE) solution concepts. 
In this regard, our work is more related to the paradigm of \textit{rational verification and synthesis} \cite{AbateGHHKNPSW21,fisman2010rational}, where a given (temporal logic) property is checked within the set of system equilibria, or equilibrium strategies (plans) are synthesised. There are works in this research strand in the domain of probabilistic systems, e.g., \cite{kwiatkowska2021automatic,gutierrez2021rational,mittelmann2023formal,hyland2024rational}. However, to the best of our knowledge, no work in this research line incorporates causal responsibility.

From the perspective of logical specification language, we mention the work that uses logical framework to reason about responsibility, e.g., \cite{yazdanpanah2019strategic,NaumovT21,naumov2019blameworthiness,ciuni2009attributing}. All of these works differ from ours in the model (deterministic vs probabilistic, one-shot vs temporally extended), the concept of responsibility, or the research aims (c.f., attribution vs planning).

 \paragraph{Outline.}
 The rest of the paper is organised as follows: 
 Section \ref{sec:model} formalises the model based on parametric stochastic game structures with parameters and utility functions over finite traces. 
 Section \ref{sec:logic} introduces \rpatl, as an extension of PATL, incorporating quantified reward and responsibility formulae for reasoning about responsibility-aware multi-agent planning. 
 Section \ref{sec:problem} formulates the core problem addressed in this paper: identifying optimal probabilistic strategies that effectively balance causal responsibilities among agents and utility awards. The section outlines the detailed procedure using parametric model checking and game-theoretic verification techniques. 
 Section \ref{sec:concl} summarises the key contributions of the study and highlights several avenues for future research.

\section{Parametric Model of Stochastic Game}
\label{sec:model}

In this section, we introduce the model utilised in this paper. We begin with the conventional model of a \textit{concurrent stochastic multi-player game} (CSG)~\cite{shapley1953stochastic,kwiatkowska2021automatic}. Then, we define the corresponding \textit{parametric stochastic MAS}, which captures the dynamics of a given CSG.

\begin{definition}
\label{def:csg}
A \textit{concurrent stochastic multi-player game} (CSG) is a tuple $ \Game = (\AG, S, s^{0}, (\ACT_i)_{i \in \AG}, \delta, \ASP, L )$ where:
\begin{itemize}
	\item $\AG=\{1, \dots, n\}$ is a finite set of \emph{agents};
	\item $S$ is a finite non-empty set of \emph{states};
	\item $s^0 \in S$ is the \emph{initial state};
	\item $ \ACT_i $ is a finite set of actions for $ i $. With each agent $ i $ and state $ s \in S $, we associate a non-empty set $ \ACT_i(s) $ of \textit{available} actions that $ i $ can perform in $ s $. Write $ {\ACT}^{\AG} = \ACT_1 \times \cdots \times \ACT_n $.
	\item $ \delta: S \times {\ACT}^{\AG} \to \dist(S) $ is a probabilistic transition function;
	\item $\ASP$ is a finite set of \emph{atomic propositions}; 
	
	\item $L: S \FS 2^{\ASP}$ is the state labelling function mapping each state to a set of atomic proposition drawn from $\ASP$.
	
\end{itemize}
\end{definition}

Following~\cite{kwiatkowska2021automatic}, we augment CSGs with \textit{reward structures} of the form $r=(r_s, r_a)$, where $r_s: S \to \RRR$ is the state reward and $r_a: \ACT^{\AG} \to \RRR$ is the action reward, and consider \textit{cumulative rewards}, that is the sum of payoffs accumulated during the run until a specific point. 

In this work, we assume that players have memoryless strategies. Informally, a memoryless strategy for player $i$ prescribes, from each state $s$, the probability of each action $a \in \ACT_i(s)$ being chosen. The set of all memoryless strategies for player $i$ from state $s$ can be encoded by a set of variables $V^i_s = \{ x_a : a \in \ACT_i(s) \}$. Intuitively, the value of $x_a \in V^i_s$ corresponds to the probability of action $a$ being chosen by player $i$ in state $s$.
Let $ V^i = \bigcup_{s \in S} V^i_s $.
A memoryless (mixed) strategy for $ i $ in $ \Game $ thus corresponds to an \textit{evaluation} of such a set of variables represented by a function $ C^i : V^i \to \mathbb{R} $. 

\begin{definition}
\label{def:model}
Given a CSG $\Game = (\AG, S, s^0, (\ACT_i)_{i \in \AG}, \delta, \ASP, L )$, we construct the corresponding \emph{parametric stochastic multi-agent system} (\psmas) as a tuple $\MMM=(\AG, S, s^0, V, \Delta, \ASP, L)$, where:
\begin{itemize}

\item $\AG=\{1, \dots, n\}$ is a finite set of \emph{agents};

\item $S$ is a finite non-empty set of \emph{states};

\item $s^0 \in S$ is the \emph{initial state};

\item $\ACT = \{a_1, a_2, \dots, a_m \}$ is a non-empty finite set of actions;

\item $V^i = \{x_{i,1}, \dots, x_{i,m}\} \subseteq V$ is a finite set of variables (\emph{parameters}) over $\RRR^m$ for each agent $i$; 

\item $\Delta: S \times \ACT^{\AG} \times S \to \FFF_{v} $ is the \emph{probabilistic transition function}, where $\FFF_v$ is the set of polynomials over $V$ with rational coefficients which can be viewed as a parametric transition probability matrix that respects the distribution in $ \delta $;

\item $\ASP$ is a finite set of \emph{atomic propositions}; 

\item $L: S \FS 2^{\ASP}$ is the state labelling function mapping each state to a set of atomic proposition drawn from $\ASP$.

\end{itemize}
\end{definition}

We introduce the notion of \textit{admissible} evaluations in a given \psmas. An evaluation $ C^i $ is admissible if 
\begin{enumerate}
	\item $ \Delta_{C^i}(s,\alpha,s) \in [0,1] $ for all $ s,s' \in S $ and $ \alpha \in \ACT^{\AG} $, i.e., each action profile has a probability between 0 and 1;
	\item $ C^i(x) \in [0,1] $ for all $ i \in \AG $ and $ x \in V^i $, i.e., for each player, each action probability is between 0 and 1;
	\item $ \sum_{x \in V^i_s} C^i(x) = 1 $ for all $ i \in \AG $ and $ s \in S $, i.e., For each player and each state, the total of action probability values should equal 1.
\end{enumerate}

Henceforth, we assume that the model takes the form of a \psmas. When evaluating solutions, we further assume that such solutions are admissible.

\begin{example}
\label{eg:model}
\begin{figure} 
\begin{minipage}{0.95\linewidth} \centering
\scalebox{1}{
  \begin{tikzpicture}[->,auto,node distance=3.5cm, thick,main node/.style={circle,draw,font=\sffamily\bfseries}]
      \node[main node, initial] (1) {$s_0$};
      \node[main node]  [right of=1](2) {$s_1$};
      \node[main node] [below of=1](3) {$s_2$};
      \node[main node] [below of=2] (4) {$s_3$};
      \path (1) edge[bend left=20] node [above] {$p_2.\alpha_2$}(2);
      \path (1) edge[bend left=20] node [below] {$p_3.\alpha_3$}(3);
      \path (1) edge[bend left=20] node [below] {$p_4.\alpha_4$}(4);
      \path (2) edge[bend left=20] node [above] {$p_1.\alpha_1$}(1);
      \path (2) edge[bend left=20] node [below] {$p_3.\alpha_3$}(3);
      \path (2) edge[bend left=20] node [right] {$p_4.\alpha_4$}(4);
      \path (3) edge[bend left=20] node [left] {$p_1.\alpha_1$}(1);
      \path (3) edge[bend left=20] node [above] {$p_2.\alpha_2$}(2);
      \path (3) edge[bend left=20] node [below] {$p_4.\alpha_4$}(4);
      \path (4) edge[bend left=20] node [above] {$p_1.\alpha_1$}(1);
      \path (4) edge[bend left=20] node [above] {$p_2.\alpha_2$}(2);
      \path (4) edge[bend left=20] node [below] {$p_3.\alpha_3$}(3);
      \path(1) edge[loop above left] node {\scriptsize $p_1.\alpha_1$} ();
      \path(2) edge[loop above right] node {\scriptsize $p_2.\alpha_2$} () ;
      \path(3) edge[loop below left] node {\scriptsize $p_3.\alpha_3$} () ;
      \path(4) edge[loop below right] node {\scriptsize $p_4.\alpha_4$} () ; 
    \end{tikzpicture}
}
\end{minipage}
\begin{minipage}{0.98\linewidth}
{\small \begin{eqnarray*} 
\AG &=& \{A_1, A_2\} \\
\ACT &=& \{\catch, \iskip\} \\
\ASP &=& \{\caught_1, \caught_2\} \\
S &=& \{s_0, s_1, s_2, s_3\} \\
V^1 &=& \{x_{1,1}, x_{1,2}\} \triangleq \{x_1,1-x_1\} \\
V^2 &=& \{x_{2,1}, x_{2,2}\} \triangleq \{x_2,1-x_2\} \\
\end{eqnarray*}}
\end{minipage}\hfill
\caption{Example: catching balls with parametric probabilistic transitions}
\label{fig:eg}
\end{figure}   
Consider two agents ($A_1$ and $A_2$) catching a repeatedly thrown ball, as shown in Figure \ref{fig:eg}. 
Each agent can either `\catch' or `\iskip' catching, and possible states are:
\begin{eqnarray*}
s_0 &=&  \dropped \triangleq \neg \caught_1 \land \neg \caught_2 \\
s_1 &=&  \collision \triangleq \caught_1 \land \caught_2 \\
s_2 &=& \score_1 \triangleq \caught_1 \land \neg \caught_2 \\
s_3 &=& \score_2 \triangleq \neg \caught_1 \land \caught_2
\end{eqnarray*}
Let the probability of $A_1$ ($A_2$) taking action $\iskip$ is $x_{1,1}=x_{1}$ ($x_{2,1}=x_{2}$) and taking action $\catch$ is  $x_{1,2}=1-x_{1}$ ($x_{2,1}=1-x_{2}$) respectively, 
transition labels and the relevant probabilities are:
$p_1=x_{1} x_{2}$, $p_2=(1-x_{1}) (1-x_{2})$, $p_3=x_{1} (1-x_{2})$, $p_4=x_{1} (1-x_{2})$,
$\alpha_1=\iskip_1 \iskip_2$, $\alpha_2=\catch_1 \catch_2$, $\alpha_3=\catch_1 \iskip_2$, $\alpha_4=\catch_1\catch_2$. Let the reward for $A_1$ ($A_2$) taking action $\catch$ be $r^1_a(\catch)=2$ ($r^2_a(\catch)=1$) and taking action $\iskip$ be $r^1_a(\iskip)=1$ ($r^2_a(\iskip)=2$). 
\end{example}
\begin{definition}
\label{def:history}
A \emph{history} $\rho$ is a non-empty finite sequence $s_0 \alpha_0 s_1 \alpha_1 \dots s_k$ of states and joint actions, 
where $\alpha_i \in \ACT^{\AG}$ is the $i^{th}$ joint action, and
$\Delta(s_i,\alpha_i, s_{i+1})>0$. 
$\rho_s(i)$ denotes the $i^{th}$ state of $\rho$, and $\rho_{\alpha}(i)$ denotes the $i^{th}$ joint action of $\rho$. In this case, we may write $\rho_s(i)\TRANS{\rho_{\alpha}(i)} \rho_s(i+1)$.
Let $\HIST{s}{\MMM}$ denote the set of histories of $\MMM$ starting from state $s$.
\end{definition}
\begin{example}
\label{eg:history}
An example history of Example \ref{eg:model} starting from state $s_0$ can be:
$$\rho = s_0 \TRANS{\scriptsize\begin{array}{cc} (1-x_{1}).\catch_1 \\ x_{2}.\iskip_2 \end{array}} s_2 \TRANS{\scriptsize \begin{array}{cc} x_{1}.\iskip_1 \\ (1-x_{2}).\catch_2 \end{array}} s_3$$ $x.a$ denotes an agent taking action $a$ with probability  $x$.
\end{example}

\begin{definition}
\label{def:strategy}
A \textit{memoryless} (mixed)\footnote{This specific type of strategy is also called \textit{behavioural} \cite{cristau10}.} strategy for $i$ is a function from the set of states to a probability distribution over agent's set of actions $\sigma_i: S \FS \dist(\ACT_i)$.
A \emph{strategy profile} is a tuple of \emph{strategies} for a set of agents, it is denoted by $\vec{\sigma}=(\sigma_1, \sigma_2, \dots, \sigma_n)$. For a set (or \textit{coalition}) of agents $J \subseteq \AG$, write $\vec{\sigma}_J$ for $(\sigma_i)_{i \in J}$.
\end{definition}

\begin{definition}
\label{def:outcome}
A state $s$ and a strategy profile $\vec{\sigma}$ induce a set of histories denoted by $\HISTSTR{s}{\MMM}{\vec{\sigma}}$. For a history $ \rho \in \HISTSTR{s}{\MMM}{\vec{\sigma}} $, the probability of $\rho = s_0 \TRANS{\alpha_0} s_1 \dots \TRANS{\alpha_{k-1}} s_k$, with $s_0 = s$, is given by:

\[
\PPP(\rho) \triangleq \prod^{k-1}_{j=1} \left( \prod^n_{i=1} \left( \sigma_i (s_0 \TRANS{\alpha_0} s_1 \dots \TRANS{\alpha_{j-1}} s_j) (\alpha_j(i)) \right)\right)
\]

\end{definition}

\begin{definition}
\label{def:payoff}
The \emph{payoff} function defined as a map from a set of histories to a real value
$\payoff:\HIST{s}{\MMM} \to \RRR^{|Ag|}$. $\payoff^i$ denotes the payoff function of $i\in \AG$ and is defined as:
$$\payoff^i(\rho) \triangleq \left(\sum^{t-1}_{j=0} (r^i_a(\alpha_j)+r^i_s(s_j)) \cdot \Delta(s_j,\alpha_j,s_{j+1}) \right)$$
where $\rho = s_0 \TRANS{\alpha_0} s_1  \TRANS{\alpha_1} \dots  \TRANS{\alpha_{t-1}}  s_t \in \HIST{s_0}{\MMM}$.
\end{definition}
\begin{definition}
\label{def:plan}
Given a non-empty coalition of agents $J \subseteq \AG$, a strategy profile $\vec{\sigma}_J$, and a state $s \in S$, a \emph{joint plan} is a function $\pi: J \FS \HISTSTR{s}{\MMM}{\vec{\sigma}_J}$. $\HIST{s}{\pi}$ is the history consistent with $\pi$ from $s$. 
Let $\pi^{-i} = \pi^{\AG\setminus \{i\}} = (\pi^1, \dots \pi^{i-1}, \pi^{i+1}, \dots \pi^n)$ denote joint plan without $i$. 
\end{definition} 
\begin{definition}
\label{def:j-compatible}
Two joint plans $\pi_1$ and $\pi_2$ are  $\langle J \rangle$-compatible if the actions taken by coalition $J \subseteq \AG$ along the histories consistent with $\pi_1$ and $\pi_2$ are equivalent, denoted as $\pi_1 \sim_{\langle J \rangle} \pi_2$. 
$\PLAN{s}{\pi}{\langle J \rangle}$ represents the set of plans that are $\langle J \rangle$-compatible with $\pi$ starting from state $s$.
\end{definition}
\begin{example}
\label{eg:compatible}
Consider two joint plans $\pi_1,\pi_2$ in Example \ref{eg:model} when the ball is thrown twice, and which result in the following joint traces. 
\begin{eqnarray*}
\pi_1  &=& (\catch_{1} \iskip_{2}), (\iskip_{1} \catch_{2}) \\
\pi_2  &=& (\catch_{1} \catch_{2}), (\iskip_{1} \iskip_{2})
\end{eqnarray*}
Clearly, we have that $\pi_1 \sim_{\{A_1\}} \pi_2$. 
\end{example}
\begin{definition}[Causal Active Responsibility (CAR) \cite{ParkerGL23}]
\label{def:car}
Given a PSMAS $\MMM$,
we say $i$ bears CAR for outcome\footnote{Outcome is specified in \rpatl formula, presented in the next section.} $\varphi$ 
in joint plan $\pi$ at state $s$, 
if $\varphi$ holds for all possible histories consistent with $\PLAN{s}{\pi}{\langle \{i\} \rangle}$ while violated in some histories consistent with $\PLAN{s}{\pi}{\langle \AG \rangle}$. 
\end{definition}
\noindent Intuitively, $i$ takes CAR for the occurrence of the outcome $\varphi$ in $\pi$ starting at $s$ if keeping $i$'s actions fixed the other agents could not avoid the outcome by choosing different actions. Note that multiple agents can individually have CAR.
\begin{definition}[Degree of CAR]
\label{def:dcar}
Given $\MMM$, the degree of the responsibility of $i \in \AG$ bearing CAR for outcome $\varphi$ under joint plan $\pi$ at state $s$ is defined as  
the ratio of the probability of behaviours following $\PLAN{s}{\pi}{\langle \{i\} \rangle}$ and leading to $\omega$, relative to that of all possible behaviours leading to $\varphi$, if the occurrence of $\varphi$ is avoidable (\ie there exists some behaviours following $\pi$ and not leading to $\varphi$); and 0 otherwise.
\end{definition}
\begin{definition}[Causal Passive Responsibility (CPR) \cite{ParkerGL23}]
\label{def:cpr}
Given $\MMM$, 
we say $i \in \AG$ bears CPR for outcome $\varphi$ in joint plan $\pi$ at state $s$, 
if $\varphi$ holds for all possible behaviours consistent with $\pi$, 
while is violated in some behaviours consistent with $\PLAN{s}{\pi}{\langle \AG \setminus \{i\}\rangle}$.
\end{definition}
\noindent Intuitively, agent $i$ takes CPR for the occurrence of outcome $\varphi$ in $\pi$ starting at $s$ if keeping all other agents' actions fixed $i$ could avoid the outcome by choosing different actions.
\begin{definition}[Degree of CPR]
\label{def:dcpr}
Given $\MMM$, the degree of the responsibility of $i$ bearing CPR for outcome $\varphi$ under joint plan $\pi$ at state $s$ is defined as 
the ratio of the probability of the behaviours following $\PLAN{s}{\pi}{\langle \AG \setminus \{i\} \rangle}$ and not leading to $\varphi$ relative to that of all possible behaviours violating $\varphi$  if the occurrence of $\varphi$ following $\pi$ is achievable; and 0 otherwise.
\end{definition}

\section{The Logic \rpatl}
\label{sec:logic}
We introduce \rpatl, a variant of PATL that incorporates quantified reward and responsibility formulae. In this paper, we specifically consider \textit{bounded path} semantics.

\subsection{\rpatl \ over finite (bounded) paths}
\begin{definition}
\label{def:syntax}                                                         
Let $\MMM=(\AG, S, s^0, V, \Delta, \ASP, L)$.
The \emph{syntax} of \rpatl \ is made up of
\emph{state formulae} and 
\emph{path formulae} 
represented by $\phi$ and $\psi$, respectively.
\begin{GRAMMAR}
 \phi
     &::=&
  a 
     \VERTICAL
  \neg \phi
     \VERTICAL
  \phi \land \phi
     \VERTICAL
    \langle A \rangle \P_{\bowtie p} \lbrack {\psi} \rbrack
     \VERTICAL
    \langle A \rangle \R_{\bowtie q} \lbrack {\F_{\le k} \phi} \rbrack
     \\
    && 
     \VERTICAL \langle A \rangle \D_{\bowtie d} \lbrack {\CAR_{i,\pi}(\psi)} \rbrack
    \VERTICAL
    \langle A \rangle \D_{\bowtie d} \lbrack {\CPR_{i,\pi}(\psi)} \rbrack
      \\
  \psi
     &::=&
  \X \phi 
     \VERTICAL
  \phi  \U_{\le k} \phi 
\end{GRAMMAR}
Here $a \in \ASP$ is an \emph{atomic proposition},
$A \subseteq \AG$ is a set of agents,
$\langle A \rangle$ is the strategy quantifier,
$i \in A$ is an agent,
${\bowtie} \in \{\le, <, \ge, >\}$,
$p \in \lbrack 0,1 \rbrack$, $q, d \in \RRR$ are probability, reward and responsibility degree bounds, respectively, and $k \in \mathbb{N}$ is a time bound.  
\end{definition}
\noindent
Note that a \rpatl formula is defined relative to a state; path formula only appear within the probabilistic operator  $ \langle A \rangle\P_{\bowtie p}\,  \lbrack \cdot \rbrack$, the reward operator $\langle A \rangle \R_{\bowtie q}\, \lbrack \F_{\le k}  \phi \rbrack$, and the responsibility degree operator $ \langle A \rangle\D_{\bowtie d}\, \lbrack {\gamma_i} \rbrack$, where $\gamma_i$ denotes the responsibility operator ${\CAR_{i,\pi}(\psi)}$ or ${\CPR_{i,\pi}(\psi)}$.
The formula $\langle A \rangle \P_{\bowtie p} \ \lbrack \psi \rbrack$ expresses that the coalition $A$ has a strategy such that the probability of satisfying path formula $\psi$ is $\bowtie p$ when the strategy is followed.
The formula $\langle A \rangle \R_{\bowtie q} \ \lbrack \F_{\le k} \phi \rbrack$ expresses that the coalition $A$ has a strategy such that the expected rewards of satisfying path formula $\F_{\le k} \phi$ is $\bowtie q$ when the strategy is followed, where $\F_{\le k} \phi = \true~ \U_{\le k}~ \phi$.
The formula $\langle A \rangle \D_{\bowtie d} \ \lbrack \gamma_i \rbrack$ expresses that the degree of responsibility of $i$ of satisfying path formula $\psi$ under coalition $A$ is $\bowtie d$.

\begin{definition}
\label{def:semantics}
Given a model $\MMM$, semantics for \rpatl \  are interpreted as follows.
For a state $s \in S$ of $\MMM$, the \emph{satisfaction relation}
$s \models_{\MMM} \phi$ for state formula denotes ``$s$ satisfies $\phi$'':

\begin{itemize}

\item $s \models_{\MMM} a$ iff $a \in L(s)$.

\item $s \models_{\MMM} \neg\phi$ iff $s \not \models_{\MMM} \phi$.

\item $s \models_{\MMM} \phi \land \phi'$ iff 
  	$s \models_{\MMM} \phi$ and  $s \models_{\MMM}  \phi'$.

\item $s \models_{\MMM}  \langle A \rangle \P_{\bowtie p} \lbrack {\psi} \rbrack$ iff there exists a strategy profile of coalition $A$ such that the probability of paths from state $s$ which are consistent with the strategy and satisfy $\psi$ is $\bowtie p$, \ie 
\[\exists \vec{\sigma}_A. \PPP ( \{ \rho \in \HISTSTR{s}{\MMM}{\vec{\sigma}} \mid \rho \models_{\MMM} \psi \} ) \bowtie p.\]

\item $s \models_{\MMM}  \langle A \rangle \R_{\bowtie q} \lbrack \F_{\le k} \phi \rbrack$ iff there exists a strategy profile of coalition $A$ such that the expected accumulated reward of paths from state $s$ which are consistent with the strategy and satisfy $\psi$ is $\bowtie q$, \ie 
\[\exists \vec{\sigma}_A. \EEE^{\vec{\sigma}_A}_{\MMM} \big(s, \rew(r, \F_{\le k} \phi) \big) \bowtie q.\]
where
\[
\rew(r, \F_{\le k} \phi) (\rho) = \left \lbrace 
\begin{array}{l}
\infty \qquad \qquad \text{if} \ \forall j \le k. \rho_s(j) \not\models_{\MMM} \phi \\
\sum^{k_\phi}_{i=0} (r_a(\rho_a(i)) + r_s(\rho_s(i))) \ \text{otherwise}
\end{array}
\right.
\]
where $k_{\phi} = \min\{j-1 \mid \rho_s(j) \models_{\MMM} \phi\}$, and $k$ denotes the time bound of $\F_{\le k} \phi = \true \U_{\le k} \phi$.
 
\item $s \models_{\MMM} \langle A \rangle \D_{\bowtie d} \lbrack \CAR_{i,\pi} (\psi) \rbrack$ iff 
$\DDD^{A} (s, \CAR_{i,\pi} (\psi)) \bowtie d$
where: \\

$
\DDD^{A} (s, \CAR_{i,\pi} (\psi)) = \\
\left\{ 
\begin{array}{l} 
0 \qquad \text{if} \not\exists \pi'' \in \PLAN{s}{\pi}{\langle \AG \rangle} .
    (\forall \vec{\sigma}_A'' \in \pi''(A). \\
   \qquad \qquad \qquad \qquad 
   (\forall \rho \in \HISTSTR{s}{\MMM}{\vec{\sigma}_A''}.\rho \not\models_{\MMM} \psi))\\
\tfrac
{\PPP(\{\rho \in \HISTSTR{s}{\MMM}{\vec{\sigma}_A'} \mid \rho \models_{\MMM} \psi, \ \vec{\sigma}'_A \in \pi'(A), \ \pi' \in \PLAN{s}{\pi}{\langle \{i\} \rangle}\} )}
{\PPP(\{\rho \in \HIST{s}{\MMM} \mid \rho \models_{\MMM} \psi\})} 
\ \text{otherwise}
\end{array}
\right.
$\\

\item $s \models_{\MMM} \langle A \rangle \D_{\bowtie d} \lbrack \CPR_{i,\pi} (\psi) \rbrack$ iff 
$\DDD^{A} (s, \CPR_{i,\pi} (\psi)) \bowtie d$
where: \\

$
\DDD^{A} (s, \CPR_{i,\pi} (\psi)) =\\
\left\{
\begin{array}{l} 
0 
\qquad
\text{if} \not\exists \pi' \in \PLAN{s}{\MMM}{\langle \AG \rangle}.
    (\forall \vec{\sigma}'_A \in \pi'(A). \\
    \qquad \qquad \qquad \qquad 
    (\forall \rho \in \HISTSTR{s}{\MMM}{\vec{\sigma}'_A}. \rho \models_{\MMM} \psi) )\\
\tfrac{\PPP(\{\rho \in \HISTSTR{s}{\MMM}{\vec{\sigma}''_A} \mid \rho \not\models_{\MMM} \psi, \ \vec{\sigma}''_A \in \pi''(A) , \ \pi'' \in \PLAN{s}{\pi}{\langle \AG \setminus i \rangle}\} )} 
{\PPP(\{\rho \in \HIST{s}{\MMM} \mid \rho \not\models_{\MMM} \psi\})} 
\
\text{otherwise}
\end{array}
\right.
$

\end{itemize}

For a history $\rho \in \HIST{s_0}{\MMM}$, the \emph{satisfaction relation}
$\rho \models_{\MMM} \psi$ for a path formula $\psi$ denotes that ``$\rho$ satisfies $\psi$'':

\begin{itemize} 

\item $\rho \models_{\MMM} \X \phi$ iff $\rho_s(1) \models_{\MMM} \phi$.
  
\item $\rho \models_{\MMM} \phi \U_{\le k} \phi'$ iff there exists $i \le k$ such that:
  $\rho_s(i) \models_{\MMM} \phi'$, and $\rho_s(j) \models_{\MMM} \phi$  for all $j < i$.

\end{itemize}

\end{definition}

\begin{example}
\label{eg:car}
Consider the outcome expressed by $$\varphi = \langle A_1, A_2 \rangle \X (\dropped \lor \score_2)$$ in Example~\ref{eg:model}, 
assume agent $A_1$ ($A_2$) executes action $\iskip$ \ with probability $x_1$ ($x_2$), and $\catch$ \ with probability $1-x_1$ ($1-x_2$), and consider joint plan $\pi = (\iskip_1 \iskip_2)$ and agent $A_1$.
Note that $\forall \pi' \in \PLAN{s_0}{\pi}{\langle \{A_1\} \rangle}$, $\HIST{s_0}{\pi'} \models_{\MMM} \varphi$, (\ie for $\pi' = (\iskip_1 \catch_2)$ or $\pi'=(\iskip_1 \iskip_2)$). Therefore, while keeping the initial state and actions of $A_1$ fixed, the other agents (\ie $A_2$ in this example) could not have acted differently to avoid the occurrence of $\varphi$. Now consider $\pi''=(\catch_1 \iskip_2)$, and note that $\HIST{s}{\pi''} \not\models_{\MMM} \varphi$, \ie there exists a joint plan $\pi''\in \PLAN{s_0}{\MMM}{\langle \{A_1,A_2\} \rangle}$ which avoids the occurrence of $\varphi$. Thus, $s_0 \models_{\MMM} \CAR_{\MMM}(A_1,\pi,\varphi)$ and $A_1$ bears CAR  for $\varphi$ in $\pi$ at $s_0$,
and clearly: 
\begin{eqnarray*}
&&\PPP(\varphi) = x_1 x_2+x_1 (1-x_2)=x_1 \\
&&\PPP(\neg\varphi) = (1-x_1)(1-x_2)+x_2 (1-x_1) = 1-x_1  \\
&&\DDD^{\langle A_1,A_2\rangle}  \lbrack \CAR_{A_1, \pi}(\varphi) \rbrack = \frac{x_1 x_2+x_1 (1-x_2)}{x_1 x_2+x_1 (1-x_2)} \cdot 1 = 1
\end{eqnarray*}
\end{example}

\begin{example}
\label{eg:cpr}
Consider the outcome expressed by $$\varphi =\langle A_1, A_2 \rangle \X \collision$$ in Example~\ref{eg:model}, 
assume each agent executes $\iskip$ \ and $\catch$ \ with probability $x$ and $1-x$ respectively.
given a joint plan $\pi = (\catch_1 \catch_2)$ and agent $A_1$. Here, $\HIST{s_0}{\pi} \models_{\MMM} \varphi$.
Consider $\pi' = (\iskip_1 \catch_2)$, and 
note that $\pi' \in \PLAN{s_0}{\pi}{\langle \{A_2\} \rangle}$ and $\HIST{s_0}{\pi'} \not\models_{\MMM} \varphi$,
\ie by keeping the initial state and the actions of all other agents fixed, $A_1$ can instead execute action $\catch_1$ to  avoid the realisation of $\varphi$. Thus, $A_1$ bears CPR  for $\varphi$ in $\pi$ at $s_0$.
Clearly, 
\begin{eqnarray*}
&&\PPP(\varphi) = (1-x)^2 \\
&&\DDD^{\langle A_1,A_2\rangle} \lbrack \CPR_{A_1, \pi} (\varphi) \rbrack = \tfrac{x(1-x)}{x^2 + x(1-x) + (1-x)^2}= \tfrac{x(1-x)}{1-x+x^2}
\end{eqnarray*}
\end{example}

\subsection{Model checking \rpatl}
As \rpatl \ extends PATL, which operates within a branching-time logic framework, the fundamental model checking algorithm shares a basic structure similar to those used in CTL (see \cite{BaierK08} for details). This algorithm operates through the recursive computation of the set $\SAT(\phi)$, which represents states satisfying the formula $\phi$ within the model. Given a state formula $\phi$, the algorithm recursively evaluates the truth values of its subformulae $\phi'$ across all states, starting from the propositional formulae of $\phi$ and following the recursive definitions of each modality. The problem of model checking a \psmas \wrt arithmetic term comparison (excluding responsibility operators introduced in this paper) is similar to PATL and rPATL (see \cite{ChenL07,kwiatkowska2021automatic} for details). 
Therefore, we focus on evaluating responsibility degree formulae $\E_{s,\vec{\sigma}_A}(\D \lbrack \gamma_i \rbrack)$.

\paragraph{CAR formulae.}
If $\gamma_i = \CAR_{i,\pi}(\psi)$, we compute the probability of behaviours leading to the path formula $\psi$ when following the strategies devised by $i$ within the joint plan $\pi$, and the probability of all possible behaviours satisfying $\psi$. The degree of the CAR formula is computed as the ratio of these probabilities. Algorithm \ref{alg:dcar} outlines the procedure.
\begin{algorithm}[h!]
\caption{Calculate $\E_{s,\vec{\sigma_A}} (\D \lbrack \CAR_{i,\pi}(\psi) \rbrack)$}
\begin{algorithmic}[1]
\State \textbf{Input:} $\MMM, s, i, \pi, \psi, A$
\State \textbf{Output:} $\E_{s,\vec{\sigma_A}} (\D \lbrack \CAR_{i,\pi}(\psi) \rbrack)$ 
with parameters of $\MMM$
    \State $\kappa \gets 0$
    \State $\LLL_{\psi}, \LLL^+, \LLL^- \gets \{\}, \{\}, \{\}$

    \For{each history $\rho \in \HIST{s}{\MMM}$}
        \If {$\rho \models_{\MMM} \psi$}
             \State $\LLL_{\psi} \gets \LLL_{\psi} \cup \{\rho\}$
        \EndIf
    \EndFor
    
    \For{each $\pi'$ in compatible plans $\PLAN{s}{\pi}{\langle\{i\}\rangle}$}
        \For{each consistent history $\rho \in \HIST{s}{\pi'}$}
            \If{$\rho \models_{\MMM} \psi$}
                \State $\LLL^+ \gets \LLL^+ \cup \{\rho\}$
            \EndIf
        \EndFor
    \EndFor
    
    \For{each $\pi''$ in all possible plans $\PLAN{s}{\MMM}{A}$}
        \For{each consistent history $\rho \in \HIST{s}{\pi''}$}
            \If{$\rho \not\models_{\MMM} \psi$}                
                \State $\LLL^- \gets \LLL^- \cup \{\rho\}$
            \EndIf
        \EndFor
    \EndFor

    \If{$|\LLL^-| > 0$} \State $\kappa \gets 1$ \EndIf

    \State \textbf{Return} $p \gets \PPP(\LLL^+)/\PPP(\LLL_{\psi}) * \kappa$ 

\end{algorithmic}
\label{alg:dcar}
\end{algorithm}

\paragraph{CPR formulae.}
If $\gamma_i = \CPR_{i,\pi}(\psi)$, we compute the probability of behaviours violating the path formula $\psi$ where agent $i$ varies its actions while other agents maintain compatible actions in $\pi$. We also compute the probability of all possible behaviours violating $\psi$. The degree of the CPR formula is computed as the ratio of these probabilities. Algorithm \ref{alg:dcpr} outlines the procedure.
\begin{algorithm}[h!]
\caption{Calculate $\E_{s,\vec{\sigma_A}} (\D\lbrack \CPR_{i,\pi} (\varphi) \rbrack)$}
\begin{algorithmic}[1]
\State \textbf{Input:} $\MMM, s, i, \pi, \varphi, A$
\State \textbf{Output:} $\E_{s,\vec{\sigma_A}} (\D \lbrack \CPR_{i,\pi}(\varphi) \rbrack)$ 
with parameters in $\MMM$
    \State $\kappa \gets 0$
    \State $\LLL_{\neg\varphi}, \LLL^+, \LLL^- \gets \{\}, \{\}, \{\}$
    
    \For{each history $\rho \in \HIST{s}{\MMM}$}
        \If{$\rho \not \models_{\MMM} \varphi$}
            \State $\LLL_{\neg\varphi} \gets \LLL_{\neg\varphi} \cup \{\rho\}$
        \EndIf
    \EndFor
    
    \For{each $\pi'$ in compatible plans $\PLAN{s}{\pi}{\langle\{A\}\rangle}$}
        \For{each consistent history $\rho \in \HIST{s}{\pi'}$}
            \If{$\rho \models_{\MMM} \varphi$}
                \State $\LLL^+ \gets \LLL^+ \cup \{\rho\}$
            \EndIf
        \EndFor
    \EndFor
    
    \For{each $\pi''$ in all possible plans $\PLAN{s}{\pi}{A \setminus \{i\}}$}
        \For{each consistent history $\rho \in \HIST{s}{\pi''}$}
            \If{$\rho \not\models_{\MMM} \varphi$}
                \State $\LLL^- \gets \LLL^- \cup \{\rho\}$
            \EndIf
        \EndFor
    \EndFor

    \If{$|\LLL^+| > 0$} \State $\kappa \gets 1$ \EndIf

    \State \textbf{Return}  $\PPP(\LLL^-)/\PPP(\LLL_{\neg\varphi}) * \kappa$ 
    
\end{algorithmic}
\label{alg:dcpr}
\end{algorithm}

Algorithms \ref{alg:dcar} and \ref{alg:dcpr} both run in polynomial space, thus overall, we obtain the following theorem.

\begin{restatable}{theorem}{rpatlmc}
	\label{theo:mc-complexity}
	Model checking \rpatl\ formula is in PSPACE.
\end{restatable}
\begin{proof}
	The proof will be in two parts. (i) First, we show that model checking \rpatl\ formula without $ \D $ operators is in PSPACE. This follows from the fact that the fragment of \rpatl\ without $ \D $ operators corresponds to the logic rPATL proposed in \cite{kwiatkowska2019equilibria}, whose model checking over CSGs is shown to be in PSPACE. (ii) We show that model checking \rpatl\ formulae of the form $ \langle A \rangle \D_{\bowtie d} \lbrack {\CAR_{i,\pi}(\psi)} \rbrack$
	or
	$\langle A \rangle \D_{\bowtie d} \lbrack {\CPR_{i,\pi}(\psi)} \rbrack $ can be done in PSPACE under the memoryless strategy assumption. Notice that in Algorithms \ref{alg:dcar} and \ref{alg:dcpr}, the  plans $ \pi', \pi'' $ and history $ \rho $ have size of at most polynomial in $ \MMM $. Furthermore, the checks in lines 6, 12, and 19 can be done in polynomial time. Thus, the algorithms run in PSPACE. As such, we obtain PSPACE model checking complexity.
\end{proof}

\section{Finding Stable Joint Plans}
\label{sec:problem}

In this section, we discuss how to compute stable joint plans. To this end, we first introduce the notion of a \textit{utility function}, which intuitively is a function that considers both reward/payoff and a responsibility degree.
\begin{definition}
	\label{def:profit}
	Given $\MMM=(\AG, S, s^0, V, \Delta, \ASP, L)$, a joint plan $\pi$, we define the \emph{payoff valuation function} of $i \in \AG$ as the expected payoff of $\HIST{s^0}{\pi}$: 
	\begin{equation} 
		\label{eq:payoff} 
		\VVV^{i}_{\pi} (s^0) \triangleq \sum_{\rho \in \HIST{s^0}{\pi}} \payoff^i(\rho)
	\end{equation}
\end{definition}
\begin{example}
	\label{eg:profit}
	The expected payoff of $\pi_1$ of $A_1$ given in Example \ref{eg:compatible} can be computed as:
	\[
	\VVV^{A_1}_{\pi_1} (s^0) = 2 \cdot (1-x_{1}) x_{2} + 1 \cdot x_{1} (1-x_{2})
	\]
\end{example}
\begin{definition}
	\label{def:resp-val}
	Given $\MMM=(\AG, S, s^0, V, \Delta, \ASP, L)$, an outcome $\varphi$, and a joint plan $\pi$, we define the \emph{responsibility valuation function} of $i \in \AG$ as the responsibility degree of $\HIST{s_0}{\pi}$: 
	\begin{equation} 
		\label{eq:er}
		\ER^{i}_{\pi} (s_0) \triangleq \DDD^{\AG} \lbrack \CAR_{i, \pi} (\varphi) \rbrack + \theta \cdot \DDD^{\AG} \lbrack \CPR_{i, \pi} (\varphi) \rbrack
	\end{equation}
	where $\theta$ is a Lagrange coefficient to adjust the weight of CAR degree and CPR degree.
\end{definition}
The equation above defines the responsibility valuation function $\ER^{i}_{\pi} (s^0) $ for agent $i$. By combining CAR and CPR with appropriate weights, the responsibility valuation function provides a measure of agent $i$'s overall responsibility in achieving the outcome $\varphi$ within the specified joint plan $\pi$ and a starting state $s^0$.
\begin{definition}
	\label{def:utility-val}
	The utility function is defined in terms of the payoff valuation and responsibility valuation as: 
	\begin{equation} 
		\label{eq:utility}
		u^i_{\pi} \triangleq \lambda (\VVV^i_{\pi}(s^0), \ER^{i}_{\pi} (s^0))
	\end{equation}

 Here, $\lambda$ is a polynomial function allowing us to weigh the importance of responsibility and gaining a reward.
\end{definition}

 A simple instantiation may be a weighted linear function of the form $\lambda=\lambda_1 \VVV^i_{\pi}(s^0) - \lambda_2 \ER^{i}_{\pi} (s^0) $. Different functions for $\lambda$ can be viewed as  different norms imposed on the agents in the system \cite{ciabattoni23}, in which case our approach can be viewed as ascribing a sanction or reward based on responsibility for achieving or violating the state of affairs associated with the norm.

Given a \psmas $ \MMM $, a \rpatl \textit{path} formula $ \varphi $, and utility function $ u^i_{\pi} $ for each $ i $, our task is to compute an NE joint plan $ \pi $ that satisfies $ \varphi $. The high-level procedure is as follows.

\begin{enumerate}

\item First we compute the set of strategy profiles satisfying the formula $ \varphi$. We apply parametric model checking techniques on $ \MMM $ to compute rational valuation functions over parameters $ V^i $ for each agent $i \in \AG$, representing strategic transition probabilities in mixed strategies over each action $a \in \ACT$ in each state $ s $.

\item Then, from this set of strategies, we restrict it to a subset of strategies that are NE. We do this by formulating a set of equations corresponding to NE, and solve them with respect to the set of strategies from step 1.

\end{enumerate}

\subsection{Nash equilibrium for responsibility and utility}

\begin{definition}
\label{def:eb}
Given $\MMM=(\AG, S, s^{0}, V, \Delta, \ASP, L)$, for each agent $i$, and $\pi^{\AG \setminus i}$, a plan $\pi^i$ is a \emph{best response} \wrt utility functions
if it is the set 
$$u\BR_i(\pi^{\AG \setminus i}) \triangleq \{\pi^i \mid \max_{\pi^i} (\lambda(\VVV^i_{\pi^i, \pi^{\AG \setminus i}}(s^0), \ER^i_{\pi^i, \pi^{\AG \setminus i}}(s^0) ))\}$$

A joint plan $\pi$ is considered a mixed \emph{Nash equilibrium} (NE) if it belongs to the best response sets $\pi \in u\BR_i(\pi^{\AG \setminus i})$ for all $i \in \AG$ in $\MMM$. 
\end{definition}

\paragraph{Computing expected payoff for reachability $\VVV$.}
The computation of the payoff valuation function, as given in (\ref{eq:payoff}), involves calculating the reachability rewards using the formula:
\[ \E^i_{s,\vec{\sigma}_A} (\langle A \rangle \R \lbrack \F_{\le k} \phi \rbrack) = 
\EEE (\payoff^i( \{\rho \in \HIST{s,\vec{\sigma}_A}{\MMM} \mid \rho \models_{\MMM}\F_{\le k} \phi\}) )\]
where $\payoff^i(\rho) =  \sum^{k_{\phi}}_{j=0} \left( r^i_a(\rho_a(j)) +r^i_s(\rho_s(j)) \right)$,
$k_{\phi} = \text{min}\{k, k'\}$ and $k'\le k$ s.t. $\rho_s(k') \models_{\MMM} \phi$.

\paragraph{Computing expected responsibility $\ER$.}
The computation of the responsibility valuation function, as given in (\ref{eq:er}), involves calculating both the CAR and CPR degrees using the formula: 
\begin{eqnarray*}
&& \E_{s,\vec{\sigma}_A} (\D \lbrack \CAR_{i,\pi}(\psi) \rbrack) \\
&=& \left\{ 
\begin{array}{l} 
0 \quad \text{if} \not\exists \pi'' \in \PLAN{s}{\pi}{\langle A \rangle}. (\forall \rho \in \HIST{s}{\pi''}.\rho \not\models_{\MMM} \psi)\\
\tfrac
{\PPP(\{\rho \in \HIST{s}{\pi'} \mid \rho \models_{\MMM} \psi \  \text{ for } \pi' \in \PLAN{s}{\pi}{\langle \{i\} \rangle}\} )}
{\PPP(\{\rho \mid \rho \models_{\MMM} \psi\})} 
\quad \text{otherwise}
\end{array}
\right.\\
&& \E_{s,\vec{\sigma}_A} ( \D \lbrack \CPR_{i,\pi}(\psi) \rbrack) \\
&=& \left\{
\begin{array}{l} 
0 
\quad 
\text{if} \not\exists \pi' \in \PLAN{s}{\MMM}{\langle A \rangle}.(\forall \rho \in \HIST{s}{\pi'}. \rho \models_{\MMM} \psi) \\
\tfrac{\PPP(\{\rho \in \HIST{s}{\pi''} \mid \rho \not\models_{\MMM} \psi \ \text{ for } \pi'' \in \PLAN{s}{\pi}{\langle A \setminus i \rangle}\} )} 
{\PPP(\{\rho \mid \rho \not\models_{\MMM} \psi\})} 
\quad 
\text{otherwise}
\end{array}
\right.
\end{eqnarray*}
which can be obtained by Algorithm \ref{alg:dcar} and \ref{alg:dcpr}.

Finally, computing the best response set $u\BR_i(\pi^{\AG \setminus i})$ corresponds to
finding the set of plans $\pi^i$ that maximise the utility. 

\paragraph{Parametric best response expressions}
When performing parametric model checking against \rpatl \ formulas to evaluate the best responses formulated above, we obtain parametric expressions representing the best response strategies.
These parametric expressions contain parameters that represent the probabilities associated with different actions or strategies of the agents involved. The values of these parameters determine the optimal strategies for each agent within the joint plan.
For example, in the case of utility-based best response, the parametric expression representing the probabilities of different actions that maximise the expected cumulative rewards for each agent. Similarly, for responsibility-based best response, the parameters may represent the probabilities associated with actions that minimise the expected responsibility degree for each agent.
By varying the values of the parameters, we can explore different scenarios and analyse how changes in the probabilities of actions impact the overall effectiveness of the joint plan.

\begin{example}
\label{eg:br-payoff}
Consider the scenario from Example \ref{eg:model}, $\varphi=\langle A_1,A_2 \rangle \F_{\le 2} (\collision \lor \dropped)$. Assume the probability of $A_1$ and $A_2$ taking action $\iskip$ is $x_1$ and $x_2$ respectively, 
and the reward value of $A_1$ taking $\iskip$ and $\catch$ is $1$ and $2$ respectively, 
while the reward value of $A_2$ taking $\iskip$ and $\catch$ is $2$ and $1$ respectively.
We compute the expected payoff of the agents w.r.t $\varphi$ as follows:
{\small \begin{align*}
\EEE(\wp^{A_1}([\![\varphi]\!])) = &  2(1-x_1)(1-x_2) + 2(1-x_1)x_2^2 x_1 \\ 
& + x_1^2 (1-x_2)x_2 + 2(1-x_1)x_2(1-x_2) \\
& + 2 x_1(1-x_2)^2(1-x_1) + x_1 x_2
\end{align*}}
And a similar calculation for $\EEE(\wp^{A_2}([\![\varphi]\!]))$.
\end{example}

\begin{example}
\label{eg:br-resp}
Continuing Example \ref{eg:br-payoff}, 
$\varphi=\langle A_1,A_2 \rangle \F_{\le 2} (\collision \lor \dropped)$,
and consider a joint plan $\pi=\{(\catch_1 \iskip_2) (\iskip_1 \iskip_2)\}$,
We have:
{\small
\begin{align*}
\PPP(\LLL_{\varphi}) &= 
1+4x_1x^2_2 - 4x^2_1 x^2_2 - x^2_1 -x^2_2 - 2x_1x_2 + 4x^2_1 x_2 
\end{align*}
\begin{align*}
\DDD^{A}_{\MMM}(s_0, \lbrack \CAR_{A_1,\pi}(\varphi) \rbrack)
&= \frac{(1-x_1)(x^2_2 x_1) + (1-x_1)(1-x_2) }{ \PPP(\LLL_{\varphi}) }\\
\DDD^{A}_{\MMM}(s_0,  \lbrack \CAR_{A_2,\pi}(\varphi) \rbrack) &=
\frac{ x_1 x_2 (x_2 - x_1 x_2 + 1) }{ \PPP(\LLL_{\varphi}) } 
\end{align*}}
\end{example}

\paragraph{Finding optimised plans by solving equilibria equations}

\begin{restatable}{theorem}{nethm}
	\label{theo:ne}
	Given $\MMM=(\AG, S, s^{0}, V, \Delta, \ASP, L)$, if the utility function $u^i$ is monotonic on $i$'s mixed strategies, then the (mixed) joint plan $\pi=(\pi_1, \dots \pi_n)$ is a mixed NE of $\MMM$ \emph{iff} for each $i \in \AG$, every pure strategy (with probability $1$) of $\pi^i$ is a best response to $\pi^{\AG \setminus i}$.
\end{restatable}
\begin{proof}
	The theorem establishes the equivalence between being a mixed Nash Equilibrium and every pure strategy being a best response to the complementary mixed strategies. The proof can be obtained by contradiction. 
	
	(``$\Rightarrow$''): if an action $a$ in the support of $\pi^i$ is not a best response of $\pi^{\AG \setminus i}$, then, due to monotonicity, reallocating probability to a best response would increase/decrease $\nu_i$ (representing valuation function for $u$, $\gamma$ or $\tau$) contradicting $\pi^i$ being a best response.
	
	(``$\Leftarrow$''): if a plan $\pi'^{i}$ gives higher/lower expected valuation than $\pi^i$ in response to $\pi^{\AG \setminus i}$, at least one action in $\pi'^{i}$ provides higher/lower valuation than some action in $\pi^i$, contradicting $\pi^i$ being a best response.
	
\end{proof}

\noindent The theorem establishes the equivalence between being a mixed Nash Equilibrium and every pure strategy being a best response to the complementary mixed strategies. The proof can be obtained by contradiction, details are provided in supplementary material. 

\Cref{theo:ne} has implications for computing NE. If a mixed plan $\pi$ is a best response, then each pure strategy within the mix must also be a best response, implying that all pure strategies in the mix yield the same expected utility. In other words, every choice in the support of any agent's equilibrium mixed plan must result in the same valuation: $u^i(\pi^{-i}, \pi^i) = u^i(\pi^{-i}, \pi^{i\prime})$ for any two actions $a, a' \in \ACT$ corresponding to $\pi^i, \pi^{i\prime}$ with positive probabilities. This fact allows us to express the NE condition as follows:
\begin{equation}\label{eq:ne-poly}
\left\lbrace
\begin{array}{l}
u^i(\pi^{-i}, \pi^i)=u^i(\pi^{-i}, \pi^{i\prime}) \quad \forall \pi^{i},\pi^{i\prime}\\ 
\sum^m_{j=1} \pi_{ij}=1 \quad \forall i \\
0 \le \pi_{ij} \le 1 \quad \forall i, j.
\end{array}
\right.
\end{equation}

Solving these equations identifies the NE and finds the optimal joint plan for given objectives. Thus, we obtain the following theorem. The proof is obtained from a reduction to the existential theory of the real numbers~\cite{canny1988some}.

\begin{restatable}{theorem}{necomplex}
\label{theo:ne-complexity}
Computing NE joint plan satisfying \rpatl formula $\varphi$ can be done in PSPACE.
\end{restatable}
\begin{proof}
	Observe that \Cref{eq:ne-poly} contains a set of polynomial equations with $|\AG|$ variables, where each equation has a degree of at most $|\AG|$. This corresponds to the problem of solving a system of polynomial equations which, with the fact that the existential theory of the real numbers is decidable in PSPACE \cite{canny1988some}, can also be solved in PSPACE.
\end{proof}

\begin{example}
\label{eg:ne-ubr}
Continuing Example \ref{eg:br-payoff} (where we use a linear weighted function such that $\lambda_1=1, \lambda_2=0$) for best response calculation, by Theorem \ref{theo:ne}, we have:
$$\left\lbrace
\begin{array}{l}
2(1-x_2) + 4x_2(1-x_2) = x_2+2(1-x_2)x_2 \\
2(1-x_1)=x_1+8(1-x_1)x_2 \\
0 \le x_1 \le 1 \\
0 \le x_2 \le 2
\end{array}
\right.$$
by solving the above NE conditions, we get: 
$$x_1 = \sqrt{17}/4 - 1/4 \approx 0.78, \ x_2 = 2/3,$$
\ie when $A_1$ takes action of $\iskip$ with probability of $0.78$ and $A_2$ takes action of $\iskip$ with probability of $2/3$, an NE equilibrium is reached.
\end{example}

\begin{example}
\label{eg:ne-rbr}
Continuing Example \ref{eg:br-resp} for best response calculation, by Theorem \ref{theo:ne}, we have:
$$\left\lbrace
\begin{array}{l}
\frac{(1-0)(0 \cdot x^2_2+1-x_2)}{1+4 \cdot 0 \cdot x^2_2 - 4\cdot 0^2 \dot x^2_2 - x^2_1 -x^2_2 - 2 \cdot 0 \cdot x_2 + 4\cdot 0^2 \cdot x_2} \\
\qquad = \frac{(1-1)(1 \cdot x^2_2+1-x_2)}{1+4 \cdot 1 \cdot x^2_2 - 4 \cdot 1^2 \cdot x^2_2 - x^2_1 -x^2_2 - 2 \cdot 1 \cdot x_2 + 4 \cdot 1^2 \cdot x_2} \\
\frac{x_1 \cdot 0 (0 - x_1 \cdot 0 + 1)}{1+4x_1 \cdot 0^2 - 4x^2_1 \cdot 0^2 - x^2_1 - 0^2 - 2x_1 \cdot 0 + 4x^2_1 \cdot 0 } \\
\qquad = \frac{x_1 \cdot 1 (1 - x_1 \cdot 1 + 1)}{1+4x_1 \cdot 1^2 - 4x^2_1 \cdot 1^2 - x^2_1 - 1^2 - 2x_1 \cdot 1 + 4x^2_1 \cdot 1 }  \\
0 \le x_1 \le 1 \\
0 \le x_2 \le 1
\end{array}
\right.$$
by solving the above NE conditions, we get: 
$x_1 =0$ and $x_2 = 1$,
\ie when $A_1$ takes action of $\catch$ while $A_2$ takes action of $\iskip$ under joint plan $\pi$, an NE equilibrium is reached.
This meets our intuition, as the outcome considered in the example for responsibility analysis is $\varphi=\langle A_1,A_2 \rangle \F_{\le 2} (\collision \lor \dropped)$, to minimise their responsibility, the best option would be for one of the agents to catch the ball. Additionally, the joint plan considered in the example is $\pi=\{(\catch_1 \iskip_2) (\iskip_1 \iskip_2)\}$, therefore, the optimised plan regarding minimised responsibility would have $A_1$ catch the ball while $A_2$ skips catching.
\end{example}
\section{Conclusions and Future Work}
\label{sec:concl}

In this paper we developed an approach to multi-agent strategic reasoning with responsibility-aware agents. We introduced \rpatl, a logic that can be used to reason about causal responsibility and temporal properties. We provide an approach to model-check \rpatl formulae against a parametric model of probabilistic MAS and show that it can be done in PSPACE and is thus no more difficult than the model checking problem of the extended rPATL logic as presented in~\cite{kwiatkowska2021automatic}. We also show how to synthesise NE joint strategies/plans that satisfy some \rpatl property, which can also be done in PSPACE.

Building on this paper, several promising avenues for future work are immediately apparent. Perhaps the most obvious is to consider finite-memory strategy settings. As shown in \cite{kwiatkowska2018automated,svorevnova2016quantitative}, bounded temporal properties often require finite-memory strategies. It would also be interesting to explore how to extend our approach to a more expressive logic, such as Probabilistic Strategic Logic (PSL)~\cite{AminofKMMR19}. While this would enable us to model complex scenarios and reason about agent behavior more easily, we must also consider the associated increase in complexity. Another avenue for exploration is the following question: what can be done if no NE joint plan satisfies the required property? One approach to addressing this is to ``repair'' the system, for example by introducing explicit \textit{norms}~\cite{bulling2016norm} or modifying the reward structure using \textit{reward machines}~\cite{ijcai2022p31}.

\bibliographystyle{splncs03}
\bibliography{BIB-resp}

\end{document}